%% file: main.tex
\documentclass[10pt,twocolumn,letterpaper]{article}

\usepackage[pagenumbers]{wacv} 

\input{preamble}

\definecolor{wacvblue}{rgb}{0.21,0.49,0.74}
\usepackage[pagebackref,breaklinks,colorlinks,allcolors=wacvblue]{hyperref}
\usepackage{multirow}
\def\wacvPaperID{} 
\def\confName{WACV}
\def\confYear{2027}

\title{Efficient All-in-One Weather Restoration using Spectral Harmonization}

\author{Paula Garrido-Mellado~$^{1,2}$\thanks{Corresponding Author}~, Daniel Feijoo~$^{1}$, Yuning Cui~$^{3}$, Alvaro Garcia~$^{1}$, Marcos V. Conde~$^{1}$\\
{\normalsize $^{1}$ Cidaut AI, Fundación Cidaut, Spain}\\
{\normalsize $^{2}$ University of Valladolid, Spain}\\
{\normalsize $^{3}$ Technical University of Munich, Germany}\\
{\tt\small\{paugar, marcos.conde\}@cidaut.es}\\
{\tt\normalsize {\href{https://github.com/cidautai/FReSH-IR}{github/cidautai/FReSH-IR}}}
 }

\begin{document}
\maketitle
\input{sec/0_abstract}    
\input{sec/1_intro}
\input{sec/2_related}
\input{sec/3_method}
\input{sec/4_experiments}
\input{sec/5_discuss}

{
    \small
    \bibliographystyle{ieeenat_fullname}
    \bibliography{main}
}

\newpage
\appendix
\input{sec/supplementary}

\end{document}

%% file: preamble.tex
\usepackage{adjustbox}
\usepackage{colortbl}
\usepackage{comment}

\definecolor{Gray}{gray}{0.9}
\definecolor{lgray}{gray}{0.95}
\definecolor{LightCyan}{rgb}{0.92,0.92,1}
\definecolor{lyellow}{rgb}{1,1,0.92}
\definecolor{lgreen}{rgb}{0.92,1,0.95}
\definecolor{dgreen}{rgb}{0.,0.6,0.}
\definecolor{dred}{rgb}{0.6,0.,0.}
\definecolor{llblue}{rgb}{0.92,0.93,0.95}
\definecolor{lred}{rgb}{1,0.85, 0.85}
\definecolor{tabhighlight}{HTML}{e5e5e5}

%% file: sec/0_abstract.tex
\begin{abstract}
Adverse weather conditions such as rain, haze, and snow significantly degrade image quality, posing challenges for both human perception and physical AI. Existing restoration methods require large computational budgets, struggling to process high-resolution images and  handle different degradations. In this paper, we present \textbf{F}requency \textbf{Re}construction via \textbf{S}pectral \textbf{H}armonization, a novel lightweight all-in-one restoration method that explicitly decomposes feature representations into high- and low-frequency components at each scale of a hierarchical encoder-decoder architecture. By combining spectral decomposition with spatial processing through Fourier-based skip connections, FReSH-IR captures complementary frequency information without sacrificing spatial detail. Our approach achieves similar restoration quality with 80\% fewer parameters and operations than transformer-based models. Extensive experiments demonstrate that our method offers a great efficiency-performance trade-off, highlighting its practical applications in constrained-resource systems.

\end{abstract}

%% file: sec/1_intro.tex
\section{Introduction}
\label{sec:intro}

\begin{figure}[t]
\centering
\begin{tabular}{c}
    \includegraphics[width=0.9\linewidth]{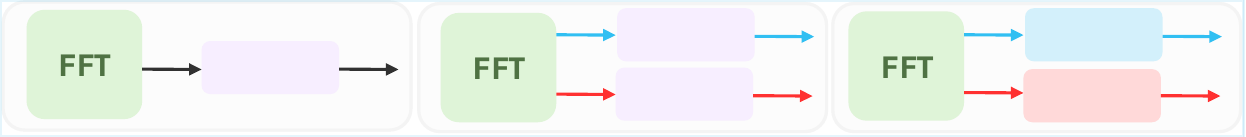} \\
    \makebox[0.9\linewidth]{\hspace{10pt}(a) \hspace{55pt} (b)\hspace{55pt} (c)}  \\
    \\
    \includegraphics[width=0.9\linewidth]{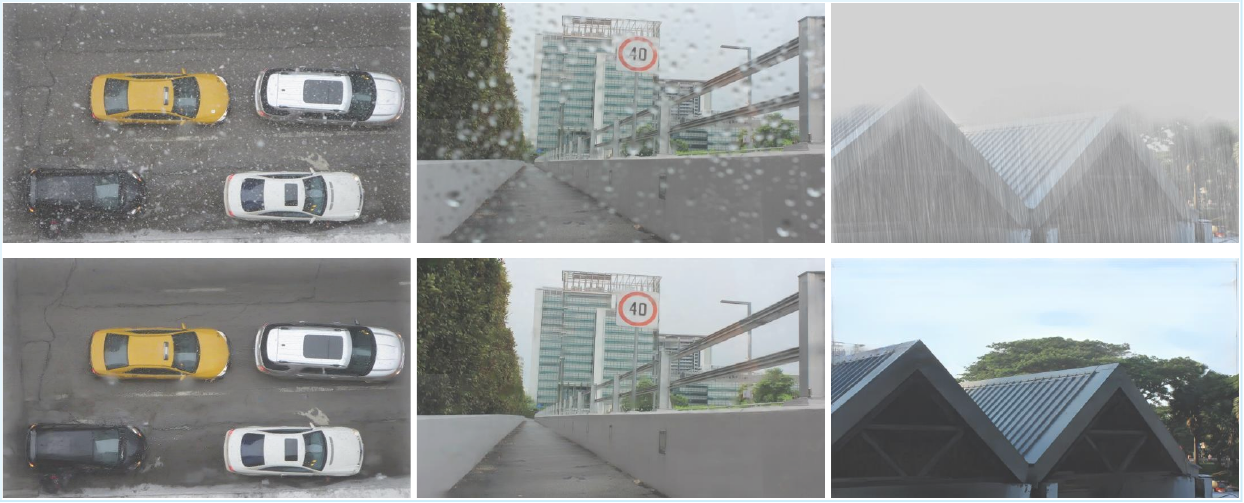} \\
\end{tabular}
\caption{(\emph{Top row}) Different approaches for frequency-domain image restoration. a) \textbf{Global spectral filtering}, \ie convert to frequency domain and process the whole spectrum without separating high/low frequencies~\cite{jiang2024sfhformer}. b) \textbf{Split LF/HF, shared processing}: separate high and low frequency components but process them using the same blocks~\cite{cui2025adair}. c) \textbf{Our approach: split LF/HF, specialized processing}: separate and process high and low frequencies using \emph{specialized} blocks for each. (\emph{Bottom row}) Restored real-world and synthetic images using our method, FReSH-IR.}
\end{figure}

Adverse weather conditions, such as snow, rain, raindrops, and haze, severely degrade visual information, making image restoration an essential task for various applications. To address this issue, researchers have explored methods to remove weather-induced degradations from images~\cite{valanarasu2022transweather}. In particular, significant efforts have been made to leverage image-based priors, such as edges and Fourier frequency patterns, as a foundation for restoring degraded images~\cite{kang2011automatic, Li_2016_CVPR, li2018superres}. 

Early research on adverse weather restoration primarily focused on handling \emph{single weather degradations}, resulting in models designed specifically for dehazing, deraining, or desnowing. However, this approach is limited in real-world scenarios where multiple weather degradations occur simultaneously, such as the concurrent presence of rain and haze. Moreover, storing multiple models and dynamically selecting the appropriate one is impractical. Despite these advancements, developing a \emph{unified model} capable of handling diverse degradations or weather conditions remains a major challenge due to the highly variable nature of weather-induced degradations~\cite{li2022all, valanarasu2022transweather}. This problem is known as All-in-One image restoration (\textbf{AIO}). The complex interactions between different atmospheric conditions lead to unpredictable degradations, making it difficult to design a single framework that generalizes across diverse scenarios.

To address the challenges posed by adverse weather conditions, several studies have explored image restoration techniques in the frequency domain. AIRFormer~\cite{gao2023frequency}, a general-purpose AIO model, constructs a frequency-guided transformer encoder by incorporating wavelet-based prior information to enhance feature extraction, leveraging structured frequency priors to improve restoration quality. Fourmer~\cite{zhou2023fourmer}, similarly, employs the Fourier transform to disentangle image degradation from content components, utilizing the global nature of the Fourier domain to comprehensively process degradation patterns. AdaIR~\cite{cui2025adair} is designed to identify and mitigate degradation patterns through frequency domain analysis. 

These methods demonstrate that \emph{operating in the spectral domain} can effectively capture weather-related degradation patterns. However, existing frequency-domain restoration approaches either process the spectrum globally or apply identical operations to all components~\cite{jiang2024sfhformer, zhou2023fourmer, cui2025adair}, limiting their ability to handle spatially localized artifacts like raindrops or snowflakes. Treating them uniformly limits restoration performance and computational efficiency.

In this work, we present \textbf{FReSH-IR} (\textbf{F}requency \textbf{Re}construction via \textbf{S}pectral \textbf{H}armonization), a lightweight all-in-one (AIO) weather restoration designed to bridge the gap between restoration quality and computational efficiency. FReSH-IR employs an asymmetric UNet~\cite{ronneberger2015unet} architecture that decomposes features into high and low Fourier frequency components at each scale and processes them through \emph{specialized} blocks tailored to their characteristics. 

FReSH-IR achieves restoration quality competitive with leading AIO methods while requiring \textbf{80\% fewer parameters}, \textbf{up to 90\% fewer MACs}, and running with \textbf{5$\times$ faster inference} than top-performing approaches such as SSGFormer~\cite{jeong2025robust}. This positions FReSH-IR as the most efficient AIO weather restoration model with the best efficiency-quality trade-off. 

%% file: sec/2_related.tex
\section{Related Work}
\label{sec:related}

\subsection{Single-task Adverse Weather Removal}
The first attempts at removing adverse weather conditions from images involved single-task or task-specific models, which focused on restoring only one type of degradation.

\textbf{Raindrop Removal}. First approaches~\cite{you2015adherent} considered temporal information from videos to restore the occluded areas caused by raindrops. During the machine learning era, CNN and GAN-based approaches later improved performance by learning data-driven degradation priors~\cite{eigen2013restoring, qian2018attentive, yan2022raingan}. The authors of ~\cite{quan2019deep} showed significant advancements in this field by introducing edge information. The dataset proposed by~\cite{jin2024raindrop} comprises both day and night images, including drop-focused and background-focused images.

\textbf{Dehazing}. This task has been widely studied, from the use of CNNs~\cite{cai2016dehazenet} to more complex techniques such as Vision Transformers ~\cite{song2023vision}. In~\cite{li2017aod}, Li \etal considered atmospheric luminosity and transmission maps to restore hazy images. The authors of ~\cite{chen2024dea} included a content-guided attention mechanism, and Zhang \etal ~\cite{zhang2021hierarchical} created a hierarchical density-aware network. In ~\cite{cui2023focal}, Cui \etal proposed a strong baseline that works in both the spatial and frequency domains.

\textbf{Deraining}. Early methods relied on learning-based approaches~\cite{fu2023continual}. Working in the frequency domain~\cite{he2024dual, gao2024efficient, jiang2023dawn} has become a common and effective technique, demonstrating improved visual and quantitative results. Some studies also use Transformers ~\cite{chen2023learning, xiao2022image}, and recently, Mamba-based methods~\cite{zou2024freqmamba,li2025ms} have become more popular.

\textbf{Desnowing}. The authors of~\cite{chen2020jstasr} were among the first to work on this task, including transparency awareness in their network. Another pioneering work is~\cite{liu2018desnownet}, which focused on context-awareness and utilized CNNs. In~\cite{zhang2021deep}, the authors introduced semantic and geometric priors, and in~\cite{chen2021all}, a dual-tree wavelet transform was used. The authors of~\cite{chen2022snowformer} presented a Transformer-based method. Lastly, Lai \etal~\cite{lai2025snowmaster} used Multi-Model Optimization and Multimodal Large Language Models to solve this degradation.

\subsection{All-in-One Image Restoration}
Since the specific-task approach involves using several models to solve real-world degradations, which is computationally challenging, the use of a single model to restore all types of degradations has become a powerful research direction. Li \etal~\cite{li2020all} proposed the first AIO model for adverse weather. Numerous methodologies in this domain are based on Transformer architectures such as TransWeather, MWFormer, and others~\cite{valanarasu2022transweather, zhu2024mwformer, zeng2025all, liang2021swinir, wang2022uformer}. Additionally, some approaches incorporate frequency mining~\cite{cui2025adair}, Mixture-of-Experts~\cite{luo2023wm, wang2025moerl}, or histogram-based attention modules~\cite{sun2024restoring}. Prompt-based methods are also popular in AIO image restoration~\cite{potlapalli2023promptir, conde2024instructir, wu2025learning}. In~\cite{zhu2023Weather}, the authors proposed a new training paradigm and provided a thorough analysis of weather degradations.

%% file: sec/3_method.tex
\section{Method}
\label{sec:method}
\subsection{Preliminaries: Frequency Spectrum Analysis}
\label{subsec:preliminaries}
Recent works incorporate frequency information into their models to restore weather degradations using various techniques, such as the Sobel operator \cite{jeong2025robust}, the fast Fourier transform (FFT) \cite{cui2025adair}, and average pooling \cite{cui2023focal}. To determine the most effective approach for isolating weather-induced degradations, we conduct a systematic comparison of four representative methods for extracting high-frequency components, included in the Supp. Material.

Based on this analysis, we adopt a hybrid strategy for high-frequency selection across the different levels of the network. In early layers, where low-level degradation patterns are most salient, we employ FFT-based decomposition with stricter masks (\ie, high cutoff values) to achieve precise isolation of weather artifacts. As the network deepens, the Fourier masks are progressively relaxed. 
In the bottleneck of the architecture, we transition to GAP-based extraction, which, despite operating on deep feature representations, retains the ability to capture structural image edges. This property is particularly advantageous for the reconstruction of fine-grained image details in the later stages of the restoration process, where preserving structural information is critical for achieving faithful outputs.

\subsection{Overall Framework}
\begin{figure*}[t]
  \centering
  \includegraphics[width=\linewidth]{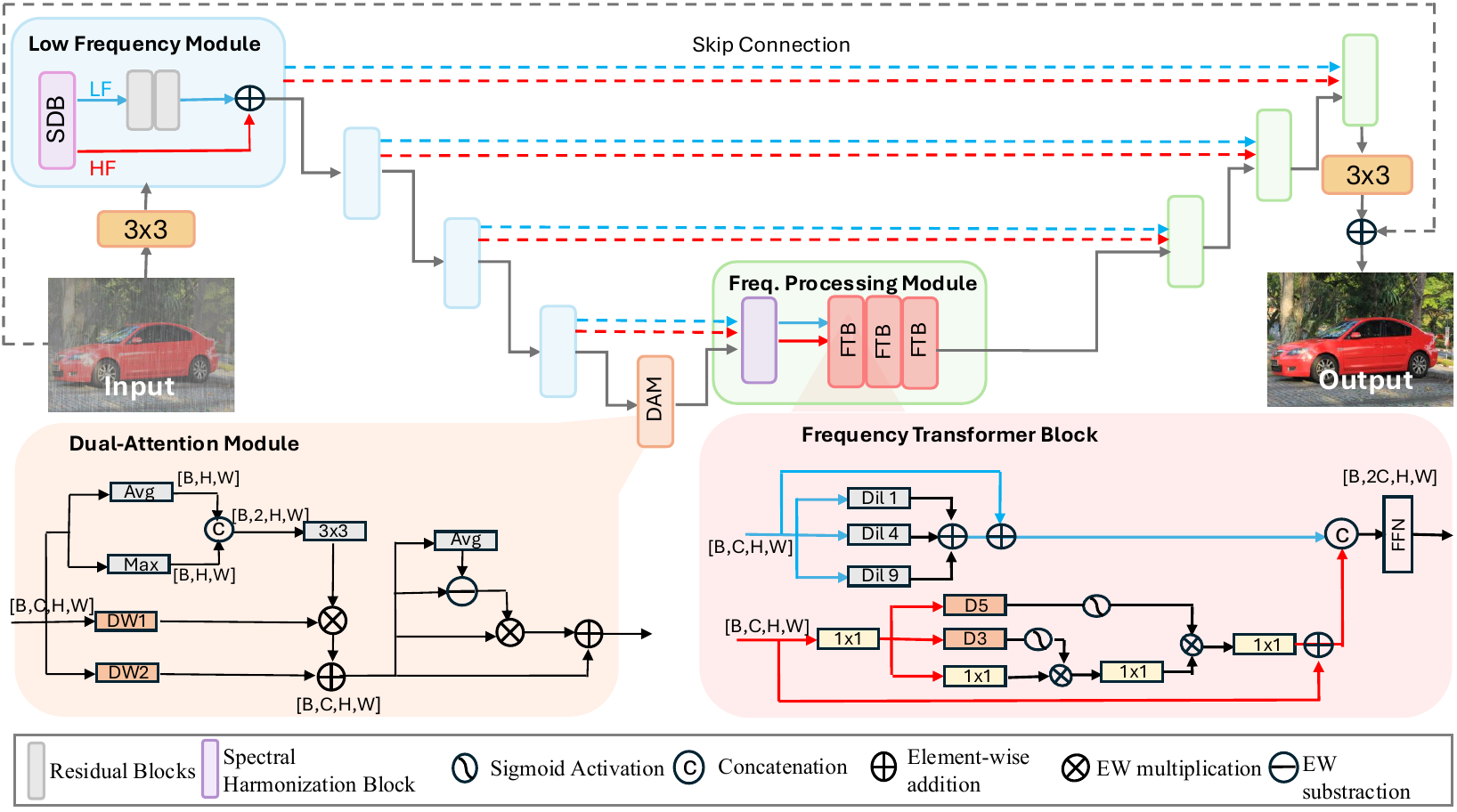}
  \caption{\textbf{FReSH architecture.} The encoder consists of a Spectral Decomposition Block (SDB) and two Residual Blocks. Each encoder block sends the \textbf{L}ow \textbf{F}requencies (blue arrows) and \textbf{H}igh \textbf{F}requencies (red arrows) to the same-level decoder block. These spectral skip connections are integrated with the restored features through the Spectral Harmonization Block and fed into the Frequency Transformer Blocks. Finally, the middle blocks integrate the Dual-Attention Module (DAM).
  }
  \vspace{-3mm}
  \label{fig:arch}
\end{figure*}

The proposed FReSH-IR follows a U-shaped CNN-based backbone, as illustrated in \Cref{fig:arch}. Given a corrupted image $I \in \mathbb{R}^{3 \times H \times W}$, where $3$ denotes the number of channels, and $H$ and $W$ denote height and width, respectively. A $3 \times 3$ convolutional layer first projects the input into shallow feature embeddings of size $ C \times H \times W$. 

These embeddings are processed through a hierarchical encoder that progressively downsamples the features while explicitly decomposing them into low- and high-frequency components at each scale. 
The low-frequency component is refined through a stack of Residual Blocks, and both components are recombined before downsampling. 

At the bottleneck, a \textbf{Dual-Attention Module (DAM)} captures long-range spatial dependencies, which,  as shown in \cite{cui2023focal}, is particularly beneficial for homogeneous degradations such as haze, which affects broad spatial regions.

A carefully designed decoder progressively recovers the features, recombining the upsampled features with the corresponding encoder features.
Each block processes the high and low components separately, only fusing them during the last block of each level.

\begin{figure*}[t]
  \centering
  \includegraphics[width=0.8\linewidth]{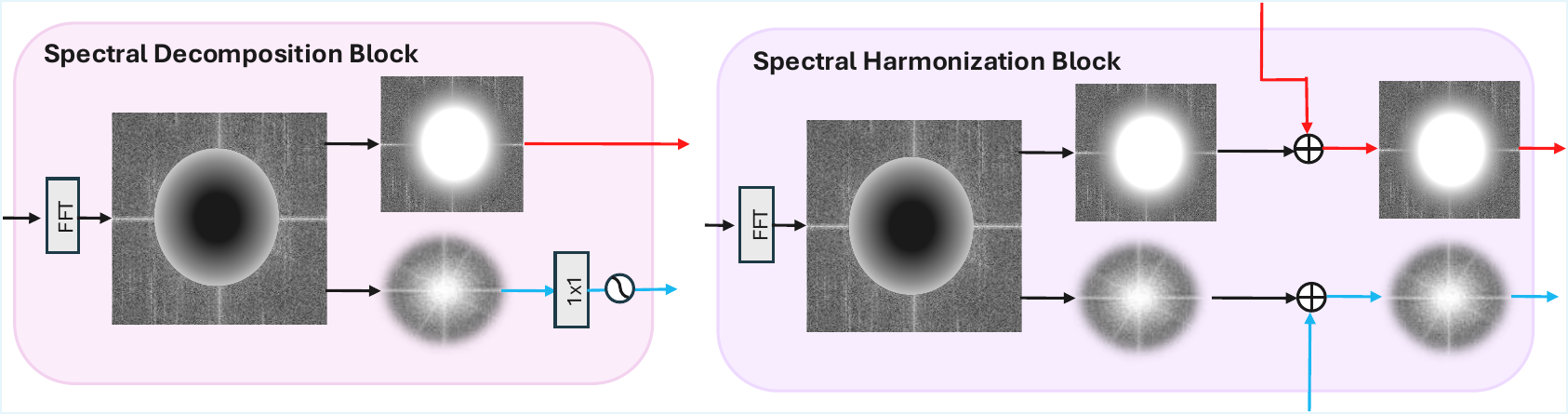}
  \caption{\textbf{(a) Spectral Decomposition Block (SDB).} The input is converted to the Fourier domain using the FFT. Applying a Gaussian mask we separate the high and low frequency components, processing the latter by a $1 \times 1$ convolution to refine them before further restoring them in the encoder blocks. \textbf{(b) Spectral Harmonization Block (SHB).} The input is divided into frequency components similar to the SDB. The computed frequencies are summed with the corresponding residual frequencies from the same-level encoder.}
  \label{fig:fourier_blocks}
\end{figure*}

\subsection{Frequency-Aware Encoder}

At each encoder level $i$, the input feature map $\mathbf{F_i}$ is decomposed into its low- and high-frequency components by the \textbf{Spectral Decomposition Block (SDB)} shown in \Cref{fig:fourier_blocks} (a), implemented via a smooth Fourier mask based on a given cutoff parameter, following:

\begin{equation}
    M(r) = \begin{cases} 
0 & r \leq r_0 - 2.5\sigma \\ 
1 - \exp\!\left(-\dfrac{(r - r_0)^2}{2\sigma^2}\right) &  r\in (r_0 \pm 2.5\sigma), \\ 
1 & r \geq r_0 + 2.5\sigma 
\end{cases}
\label{eq:mask}
\end{equation}
where $r_0 = \alpha \sqrt{H^2+W^2}$, $\sigma = 0.15r_0$ and $\alpha$ is the cutoff. 

Since the lower frequencies concentrate in the middle and the higher frequencies correspond to the spectrum edges, the higher the cutoff ratio, the larger the circle we use to eliminate frequencies from the center, thus, higher frequencies will be kept. To obtain the low-frequencies, we apply the inverse mask. We chose $\alpha = [0.4, 0.32, 0.23, 0.12]$ for each encoder level respectively.
Finally, the filtered outputs are  $\mathbf{F^{h}_i}$ and $\mathbf{F^{l}_i}$. While $\mathbf{F^{l}_i}$ captures coarse structural information, $\mathbf{F^{h}_i}$ retains fine-grained details and localized degradation patterns. 

The low-frequency component is then refined through a stack of Residual Blocks introduced in \cite{chen2022simple}, which perform lightweight attention-based feature enhancement. The refined component is $\hat{\mathbf{F}}_i^{l}$.

The encoder output at each scale is obtained by summing both components, $\mathbf{F^{enc}_i} = \hat{\mathbf{F}}_i^{l} + \mathbf{F^{h}_i}$. This design explicitly preserves high-frequency information throughout the encoding process, preventing the loss of fine structural details that are critical for weather artifact removal.

\subsection{Dual-Attention Module}

To enhance the representational capacity of the network, we propose the Dual-Attention Module (DAM), inspired by \cite{cui2023focal}, which enhances responses to important features by processing them through spatial and frequency-domain operations. 
As shown in \Cref{fig:arch}, given the output of the last encoder level, the DAM processes it sequentially.

First, we apply channel-wise average pooling and max pooling to extract the low-frequency and high-frequency components, respectively. These are concatenated and processed to produce a spatial attention map:
\begin{equation}
    \mathbf{F'_{enc}} = \operatorname{Conv}_3([\operatorname{AvgPool(\textit{F})}, \operatorname{MaxPool(\textit{F})])},
\end{equation}
where $[\cdot, \cdot]$ denotes concatenation, and  $\operatorname{Conv}_3$ is a $3 \times 3$ convolution. The resulting $\mathbf{F'_{enc}} \in \mathbb{R}^{1 \times H \times W}$ identifies key spatial locations.

Simultaneously, the input undergoes channel-wise refinement via depth-wise convolutions. One path uses cascaded $5 \times 5$ and $7 \times 7$ depth-wise dilated convolutions, with dilation factors 2 and 3 respectively (DW1). The other path employs a single $3 \times 3$ depth-wise convolution (DW2). The spatial attention map $\mathbf{F'_{enc}}$ modulates the first path:
\begin{equation}
    \mathbf{F_s} = \operatorname{DW1}(\mathbf{F})\otimes\rm{Tile}(\mathbf{F'_{enc}}, C)+\operatorname{DW2(\mathbf{F})},
\end{equation}
where $\operatorname{Tile}$ replicates $\mathbf{F'_{enc}}$ across channels. The output is a spatially enhanced feature map that we will refine by emphasizing HF components, where degraded and sharp images differ the most. At this stage, the network operates on its deepest feature representations, where GAP-based extraction proves most effective in emphasizing structural content. Therefore, we compute the HF component as:  $\mathbf{F_s^{h}} = \mathbf{F_s} -  \operatorname{Mean(\mathbf{F_s})}$.

Finally, the output of the DAM is generated by fusing these components via element-wise operations to produce the final refined feature map $\mathbf{\hat{F}}$, where perceptually important regions are emphasized.

\subsection{Spectral Skip Connections}

A key contribution of FReSH is the \textbf{Spectral Harmonization Block (SHB)}, which replaces standard additive skip connections with a frequency-aware fusion mechanism. At each decoder level, the upsampled feature map 
$\mathbf{X}$ is decomposed in the Fourier domain using the same mask used in the encoder counterpart. The decomposed components are then fused with their corresponding encoder skip features.
This frequency-aware fusion enables the decoder to selectively integrate structural LF information and HF detail from the encoder at each scale, rather than naively merging all feature channels.

\subsection{Frequency Transformer Blocks (FTBs)}
Once the feature map has been processed by the SHM, the individually fused spectral features are sent to the FTB. Each block processes the LF and HF branches independently before fusing them, which occurs in the final FTB of each level. 

The LF branch processes the features through a set of parallel dilated depthwise convolutions, composed of a regular $3 \times 3$ convolution and two dilated convolutions with factors of 4 and 9, respectively, to capture multi-scale structural information.  The HF branch is processed through a simple transformer-based block that captures long-range dependencies within fine-grained features. Both branches are then concatenated and refined through a shared FFN, similar to the FFN used in \cite{chen2022simple}. As mentioned earlier, when processing the final block of a level, a channel reduction convolution projects the fused features back to the original channel dimension. This approach allows the model to work on the separated frequencies throughout all decoder levels, while preserving the structural integrity of both components until they are fully exploited before being merged back into a unified representation and sent to the next level.

Together, the Frequency-Aware Encoder, Dual-Attention Module, Spectral Skip Connections and Frequency Transformer Blocks form a cohesive pipeline that explicitly models both structural and fine-grained spectral information at every scale, enabling robust and precise restoration.

\subsection{Regularization and Loss Functions}

We use a combination of three complementary losses, each targeting a different aspect of the restoration quality:
\begin{equation}
    \mathcal{L}_{total} = \lambda_{px}\mathcal{L}_{px} + \lambda_{enh}\mathcal{L}_{enh} + \lambda_{freq}\mathcal{L}_{freq},
\end{equation}
where $\mathcal{L}_{px}$, $\mathcal{L}_{enh}$, and $\mathcal{L}_{freq}$ are the pixel loss, enhance loss, and Fourier edge loss, respectively. $\lambda_{px}$, $\lambda_{enh}$, and $\lambda_{freq}$ are the corresponding loss weights with values $1, 1, 0.05$, respectively. These values were found empirically.

\noindent{\textbf{Pixel Loss.}}
The pixel loss provides direct supervision on the final network output by computing the L1 distance between the restored image $\hat{\mathbf{I}}$ and the ground truth $\mathbf{I}_{gt}$.

\noindent{\textbf{Enhance Loss.}}
To guide the intermediate representation at the bottleneck, the enhance loss supervises the last encoder output against a downsampled version of the ground truth $\mathbf{I}_{gt}^{\downarrow}= \phi(\mathbf{I}_{gt})$, where $\phi$ denotes nearest-neighbor downsampling by a factor of 16. It combines a pixel-level L1 term with a perceptual term computed via a pretrained VGG-19:

\begin{equation}
    \mathcal{L}_{enh} = \|\mathbf{F}_{enc} - \mathbf{I}_{gt}^{\downarrow}\|_1 + \mathcal{L}_{vgg}(\mathbf{F}_{enc}, \mathbf{I}_{gt}^{\downarrow}),
\end{equation}
where $\mathbf{F}_{enc}$ is the last encoder feature map and $\mathcal{L}_{vgg}$ measures the perceptual distance in the VGG-19 space~\cite{zhang2018perceptuallpips}.

\noindent{\textbf{Fourier Edge Loss.}}
To explicitly supervise the preservation of HF details, the Fourier edge loss applies the same high-pass spectral mask $M(r)$ from \Cref{eq:mask} to a downsampled version of the ground truth, which isolates the HF target $\mathbf{I}_{gt}^{hf}$
and the loss penalizes the deviation of the encoder's HF prediction $\mathbf{F}_{enc}^{hf}$ from this target: 

\begin{equation}
\mathcal{L}_{freq} = \|\mathbf{F}_{enc}^{hf} - \mathbf{I}_{gt}^{hf}\|_2^2.
\end{equation}

This term acts as a spectral regularizer that prevents the network from suppressing fine details during degradation removal.

Jointly, these three losses ensure that the network is supervised at the pixel, perceptual, and spectral levels simultaneously. During the first $10$ epochs, the enhance loss is not used to enforce the model to preserve critical details.

%% file: sec/4_experiments.tex
\section{Experimental Results}
\label{sec:experimental}
\subsection{Implementation Details}

For a fair comparison, we use the standard benchmark for all-in-one multi-weather restoration, following~\cite{li2020all}. It is a combination of three datasets: RainDrop \cite{qian2018attentive}, Outdoor-Rain \cite{li2019heavy}, and Snow100K \cite{liu2018desnownet}. The training set encompasses $1,069$ images from RainDrop, $9,000$ images from Outdoor-Rain, and $9,000$ from Snow100K. RainDrop comprises real raindrop images. Outdoor-Rain contains synthetic images degraded by both fog and rain streaks. Snow100K features synthetic images afflicted by snow. Similarly, we used the RainDrop test dataset \cite{qian2018attentive}, Test1 dataset from Outdoor-Rain \cite{li2019heavy}, and Snow100K-L testset \cite{liu2018desnownet} for testing.
We trained the model for 500 epochs using $384\times 384$ crops and a batch size of 16. The initial learning rate was set to $1e^{-3}$ and we used the AdamW optimizer with $\beta_1, \beta_2 = 0.9$ and a weight decay of $1e^{-3}$. 

\subsection{Quantitative Results}
In \Cref{tab:allweather}, we provide a comparative analysis of distortion metrics applied to both synthetic and real datasets. Our FReSH-IR uses $52\times$ and $10\times$ fewer parameters than the fastest solutions MWFormer~\cite{zhu2024mwformer} and TransWeather~\cite{valanarasu2022transweather}, respectively. This results in lower memory requirements, while our overall performance is superior, +1.67dB and +0.08dB. We also improve the most lightweight solution, WGWS~\cite{zhu2023Weather}, by +0.49dB and $2\times$ faster runtime. We acknowledge the work of MOERL \cite{wang2025moerl}, however we do not compare against it since the code is not available. Additionally, in \Cref{tab:realworld} we include a set of experiments in a real-world benchmark, demonstrating that we achieve competitive results also in real-world scenarios.

\begin{table*}[t]
  \centering
  \small
  \setlength{\tabcolsep}{5pt}
  \renewcommand{\arraystretch}{1.05}
  \begin{tabular}{rccccc}
    \toprule
    Method & Computational Cost & Snow100K & OutdoorData & Raindrop A & Average \\
    \midrule
    All-in-One$^{\dagger}$~\cite{li2020all} & NA & 28.33/0.882 & 24.71/0.898 & 31.12/0.927 & 28.05/0.902 \\ 

    AWRCP$^{\dagger}$~\cite{ye2023adverse} & NA & 31.92/\textbf{0.934} & 31.39/0.933 & 31.93/0.944 & 31.75/0.933 \\

    PromptIR$^{\dagger}$~\cite{potlapalli2023promptir} & 35.59/158.4/52.12 & 30.91/0.915 & 30.49/0.926 & 32.56/0.943 & 31.32/0.928 \\ 

    AdaIR$^{\dagger}$~\cite{cui2025adair} & 28.78/147.45/58.64 & 31.01/0.916 & 30.85/0.929 & 32.87/0.943 & 31.58/0.929 \\

    Restormer$^{\dagger}$~\cite{Zamir2021Restormer} & 26.13/141.24/48.90 & 30.36/0.907 & 30.03/0.922 & 32.18/0.941 & 30.86/0.923 \\ 
    
    Histoformer$^{\dagger}$~\cite{sun2024restoring} & 16.92/96.99/79.06 & \underline{32.16}/0.926 & \underline{32.08}/\underline{0.939} & \underline{33.06}/\underline{0.944} & \underline{32.43}/\underline{0.936}\\
    
    SSGFormer$^{\dagger}$~\cite{jeong2025robust} & 16.65/89.84/59.69 & \textbf{32.22}/\underline{0.927} & \textbf{32.43}/\textbf{0.941} & \textbf{33.24}/\textbf{0.949} & \textbf{32.63}/\textbf{0.939} \\

    \midrule

    \rowcolor{lyellow} WGWS$^{\dagger}$~\cite{zhu2023Weather} & \textbf{2.65}/\textbf{1.53}/25.63 & 30.16/0.901 & 29.32/0.921 & 32.38/0.938 & 30.62/0.920 \\

    \rowcolor{lyellow} MWFormer~\cite{zhu2024mwformer} & 182.81/10.49/\underline{9.43} & 30.92/0.908 & 30.27/0.912  & 31.91/0.927 & 31.03/0.916 \\
    
    \rowcolor{lyellow} TransWeather$^{\dagger}$~\cite{valanarasu2022transweather} & 38.05/\underline{6.14}/\textbf{5.23} & 29.31/0.888 & 28.83/0.900 & 30.17/0.916 & 29.44/0.901 \\
    
    \midrule
    
    \rowcolor{lgray} \textbf{FReSH-IR (Ours)} & \underline{3.46}/8.32/11.42 & 30.75/0.89 & 30.84/0.90 & 31.74/0.91 & 31.11/0.90\\
    Ours vs. SSGFormer & \textcolor{dgreen}{$\downarrow 80\%$/$\downarrow 90\%$/$\downarrow 80\%$} & \textcolor{dred}{$\downarrow 4\%$} & \textcolor{dred}{$\downarrow 5\%$} & \textcolor{dred}{$\downarrow 4.5\%$} & \textcolor{dred}{$\downarrow 4.6\%$} \\
  \bottomrule
  \end{tabular}
  \vspace{4pt}
  \caption{\textbf{Comparison of FReSH-IR with SOTA methods.} We report PSNR-Y$\uparrow$ (dB) / SSIM$\uparrow$across datasets. The values for $^{\dagger}$ methods are collected from~\cite{jeong2025robust}. We also report computational cost based on the parameters (M) / MACs (G) / runtime (ms). MACs and runtime were calculated using 256px x 256px crops. The runtime was averaged over the forward pass of 1000 iterations using an NVIDIA RTX 4090. The \textbf{bold} and \underline{underlined} represent the best and second-best results, respectively. We highlight the methods capable to process images \colorbox{lyellow}{under 30ms}. FReSH-IR offers an optimal trade-off in terms of memory, speed, and performance.
  }
  \label{tab:allweather}

\end{table*}

\begin{table}[t]
  \centering
  \begin{adjustbox}{max width=\linewidth}
  \begin{tabular}{lcccc}
    \toprule
    Method & MUSIQ$\uparrow$ & CLIPIQA+$\uparrow$ & BRISQUE$\downarrow$ & NIQE$\downarrow$ \\
    \midrule
    TransWeather~\cite{valanarasu2022transweather} & 60.538 & 0.571 & 20.019 & 3.083 \\
    Histoformer~\cite{sun2024restoring} & \textbf{62.704} & \underline{0.600} & 20.443 & \underline{2.966} \\
    MWFormer~\cite{zhu2024mwformer} & 57.084 & 0.546 & \textbf{18.733} & 3.339 \\
    WGWS~\cite{zhu2023Weather} & 60.156 & 0.573 & 20.336 & 3.045 \\
    SSGFormer~\cite{jeong2025robust} & \underline{62.278} & \textbf{0.608} & 20.134 & 2.985 \\
    \midrule
    \textbf{FReSH-IR (Ours)} & 61.034 & 0.586 & \underline{19.265} & \textbf{2.908} \\
    \bottomrule
  \end{tabular}
  \end{adjustbox}
  \caption{\textbf{Perceptual metrics} on the real-world set of Snow100K. Bold and underlined denote the best and second-best results.}
  \label{tab:realworld}
\end{table}

\subsection{Qualitative Results}

\begin{figure*}[h!]
    \centering
        \includegraphics[width=\textwidth]{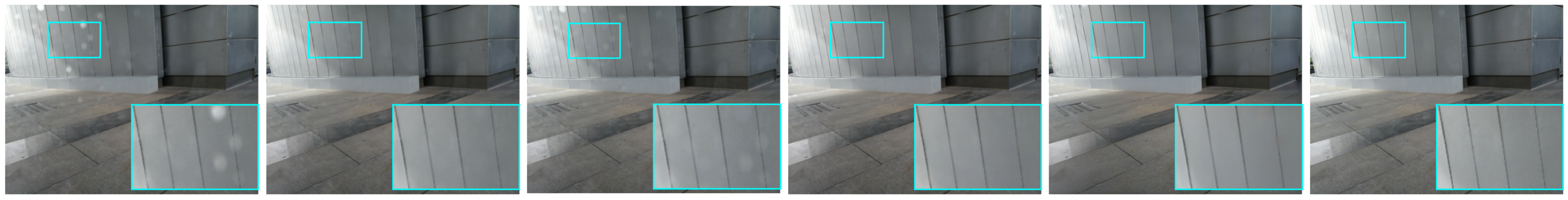}\\
        \includegraphics[width=\textwidth]{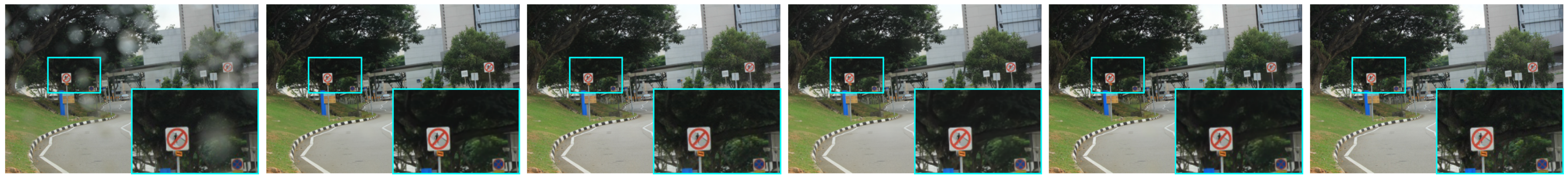}\\
    \vspace{2mm}
   
        \includegraphics[width=\textwidth]{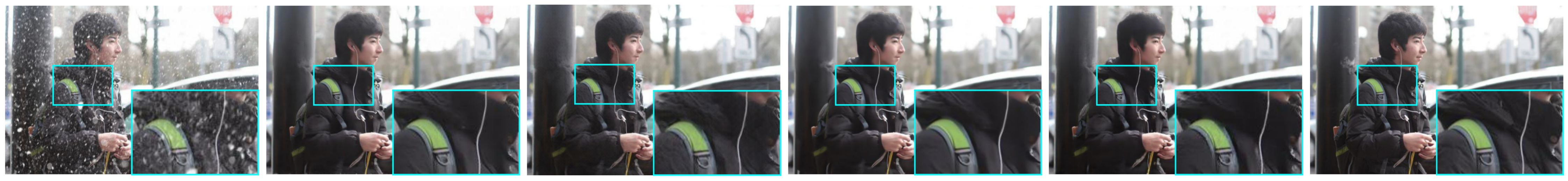}\\
   
        \includegraphics[width=\textwidth]{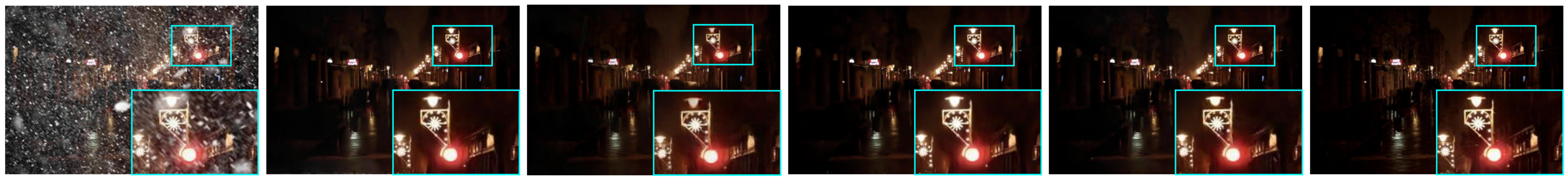}\\
    \vspace{2mm}
    
        \includegraphics[width=\textwidth]{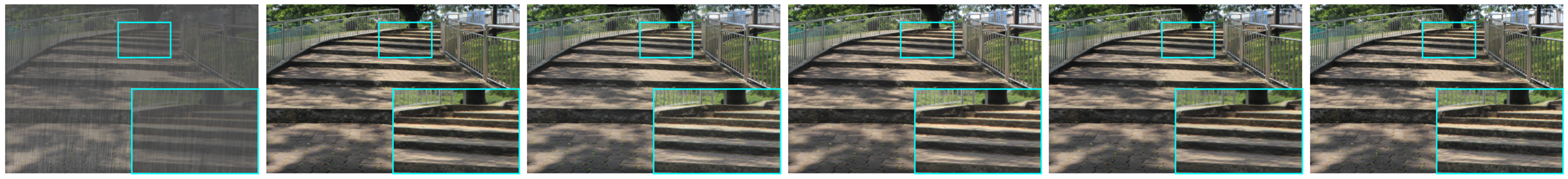}\\
        \includegraphics[width=\textwidth]{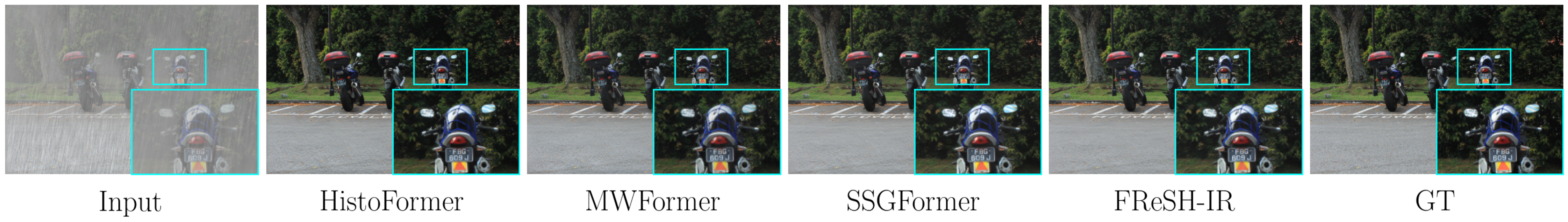}
    \caption{\textbf{Qualitative comparison of the all-in-one weather restoration methods.} Results, from top to bottom, of the following datasets: RainDrop, Snow100k and OutDoor Rain. Our \emph{FReSH-IR} provides visual results comparable to SOTA.}
    \vspace{1mm}
    \label{fig:qualis}
\end{figure*}

\begin{figure*}[h!]
    \centering
   
        \includegraphics[width=\textwidth]{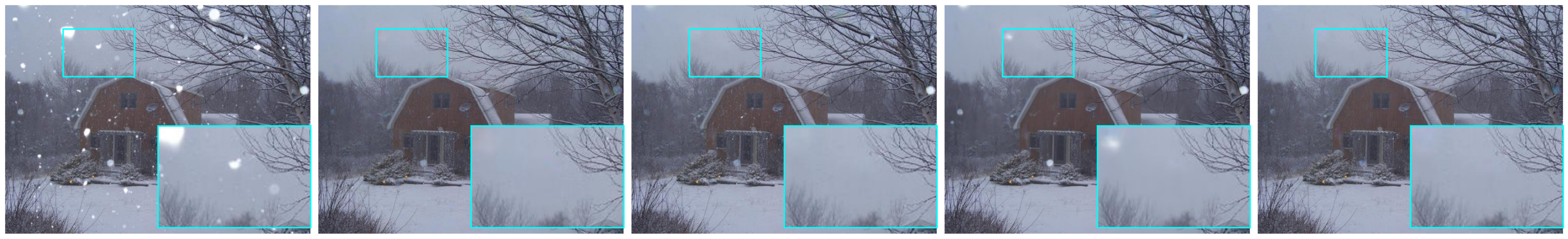}\\
   
        \includegraphics[width=\textwidth]{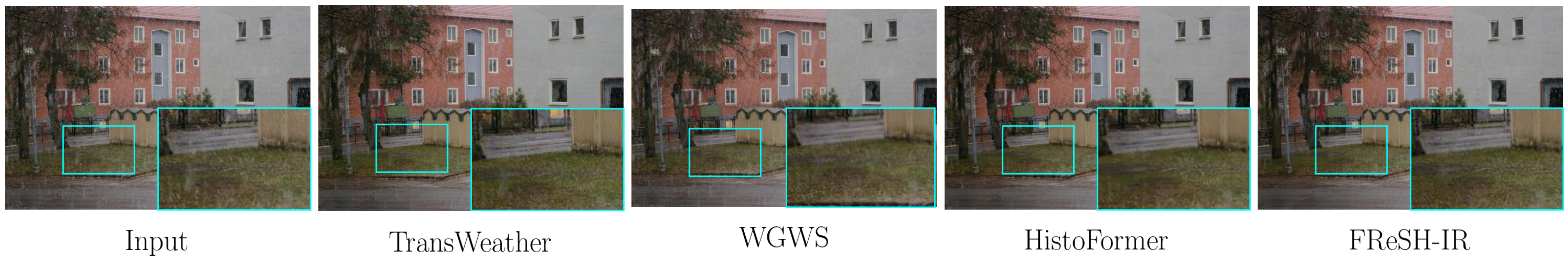}
    \caption{\textbf{Qualitative comparison of the all-in-one weather restoration methods.} Results from RealSnow dataset. Our \emph{FReSH-IR} provides visual results comparable to SOTA while being notably more efficient.}
    \vspace{-3mm}
    \label{fig:qualis_realsnow}
\end{figure*}

In this section, we show visual results on both synthetic and real-world benchmarks compared to several all-in-one weather restoration methods such as  TransWeather~\cite{valanarasu2022transweather}, WGWS~\cite{zhu2023Weather} and Histoformer~\cite{sun2024restoring}. Our visual results are comparable to SOTA methods on synthetic (\cref{fig:qualis}) and real-world benchmarks (\cref{fig:qualis_realsnow}) while saving \textbf{80\%} of computational cost against SSGFormer~\cite{jeong2025robust}. These results showcase the ability of our model to restore weather degradations while being suitable for on-device deployment. Additional qualitative results are shown in the Supplementary Material.

\subsection{Downstream tasks}
\label{sec:downstream}
In this section, we include results of the effect of our FReSH-IR on the performance of some downstream tasks such as detection and segmentation. We evaluate two segmenters covering distinct paradigms: FastSAM-s~\cite{zhao2023fast} (lightweight, class-agnostic) and SegFormer-B3~\cite{xie2021segformer} (high-capacity, semantic), and YOLO11 \cite{yolo11_ultralytics} for object detection. In \Cref{tab:downstream} we show the improvement of these tasks performance across the Rainy Cityscapes dataset, where applying our restoration method as a preprocessing step yields consistent improvements over the degraded input.

\begin{figure}[t]
    \centering
    \includegraphics[width=0.5\textwidth]{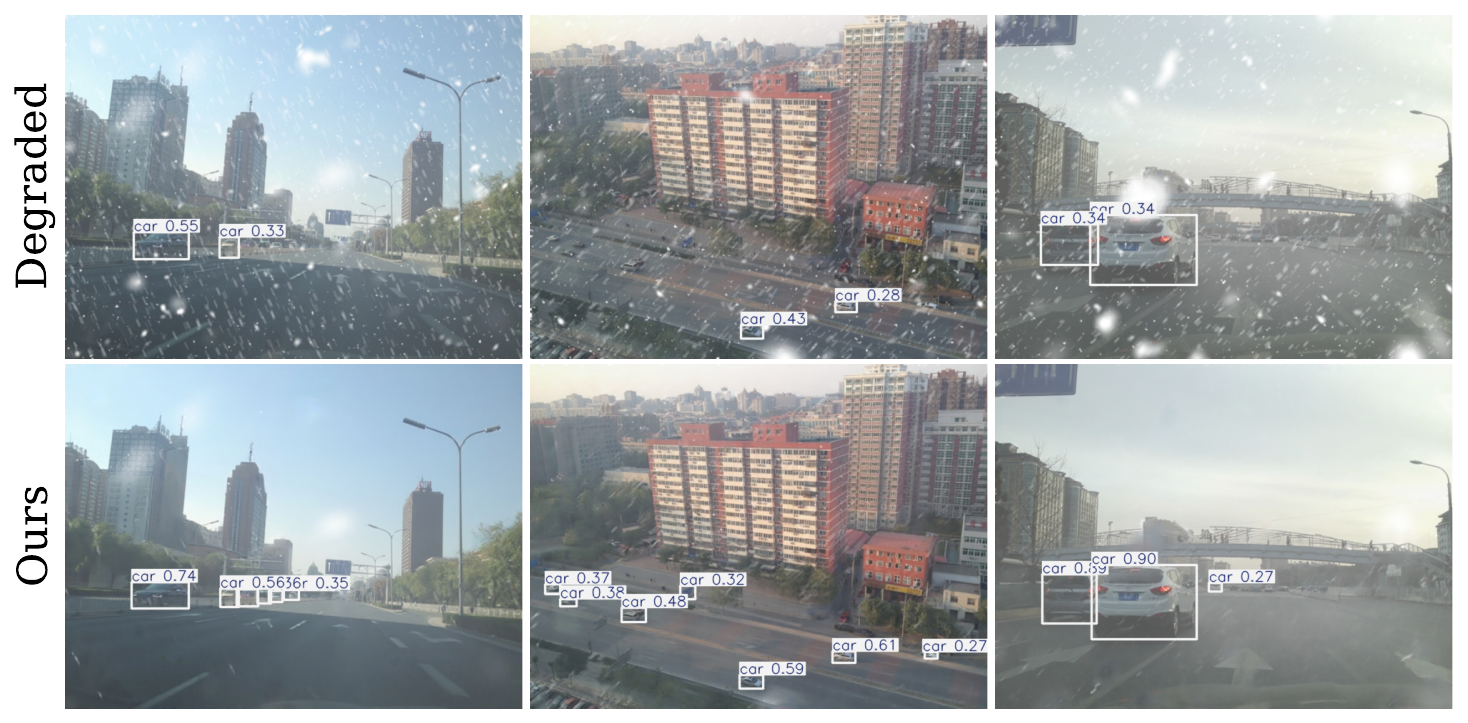}
    \vspace{1mm}
    \includegraphics[width=0.5\textwidth]{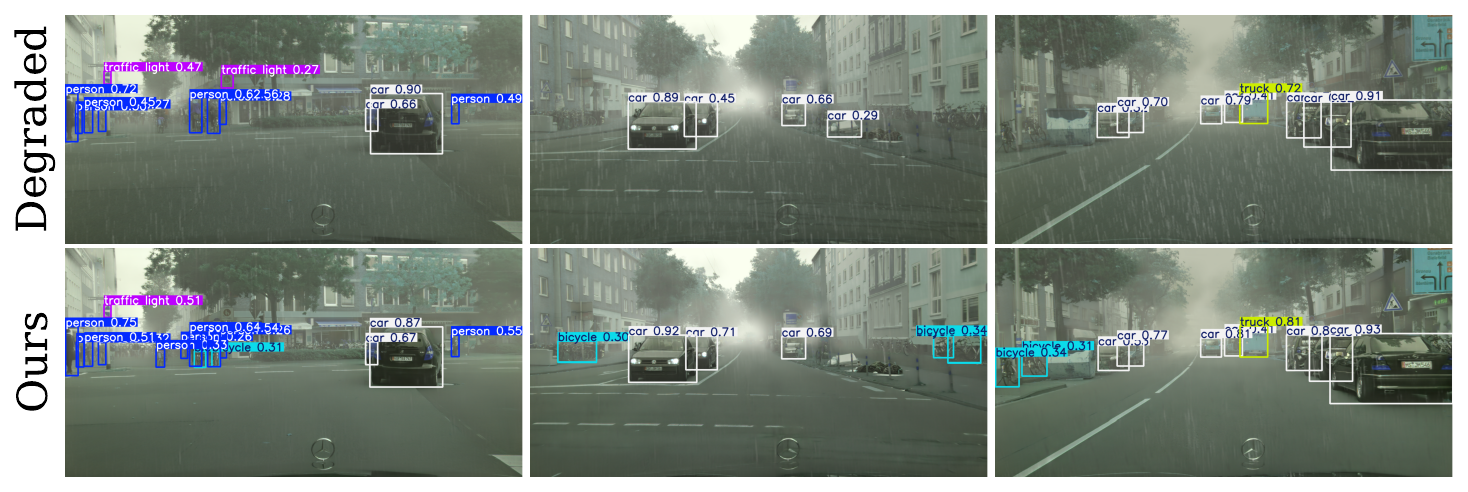}
    \caption{\textbf{Qualitative comparison on downstream tasks.} Results from object detection on the out-of-distribution CSD (top) and Rainy Cityscapes datasets (bottom), before and after restoration. Our method recovers fine details that enable the detector to recognize objects missed in the degraded inputs.}
    \label{fig:detection}
\end{figure}

We also provide qualitative results on detection in \Cref{fig:detection}. By comparing the detector's predictions on degraded images against those obtained after restoring them with our method, we observe clear improvements in both detection confidence and the number of correctly identified objects.

These experiments confirm that \emph{FReSH-IR not only restores visual quality but also yields measurable benefits for downstream perception tasks}, particularly in safety-critical scenarios such as autonomous driving.

\begin{table}[t]
  \centering
  \begin{adjustbox}{max width=\linewidth}
  \begin{tabular}{llcccc}
    \toprule
    Task & Model / Metric & Input & Restored & Clean & Rec. Gap \\
    \midrule
    \multirow{2}{*}{Segmentation}
      & SegFormer-B3      & 0.531 & 0.551 & 0.639 & 18\%   \\
      & FastSAM           & 0.277 & 0.302 & 0.343 & 37.3\% \\
    \midrule
    \multirow{3}{*}{Detection}
      & mAP$[0.5{:}0.95]$ & 0.163 & 0.176 & 0.226 & 21\% \\
      & mAP$0.5$          & 0.261 & 0.288 & 0.361 & 26\% \\
      & mAP$0.75$         & 0.165 & 0.177 & 0.229 & 19\% \\
    \bottomrule
  \end{tabular}
  \end{adjustbox}
  \caption{\textbf{Downstream task performance} before and after restoring the images with FReSH-IR on Rainy Cityscapes.}
  \label{tab:downstream}
\end{table}

%% file: sec/5_discuss.tex
\section{Discussion}
\label{sec:discuss}
\Cref{sec:experimental} results' demonstrate the performance capabilities of our method. To substantiate these strengths, we provide a series of discussions on the model design and efficiency.

\subsection{Efficiency Concerns}
\label{subsec:efficiency}
\begin{figure}[t]
  \centering
  \includegraphics[width=\linewidth]{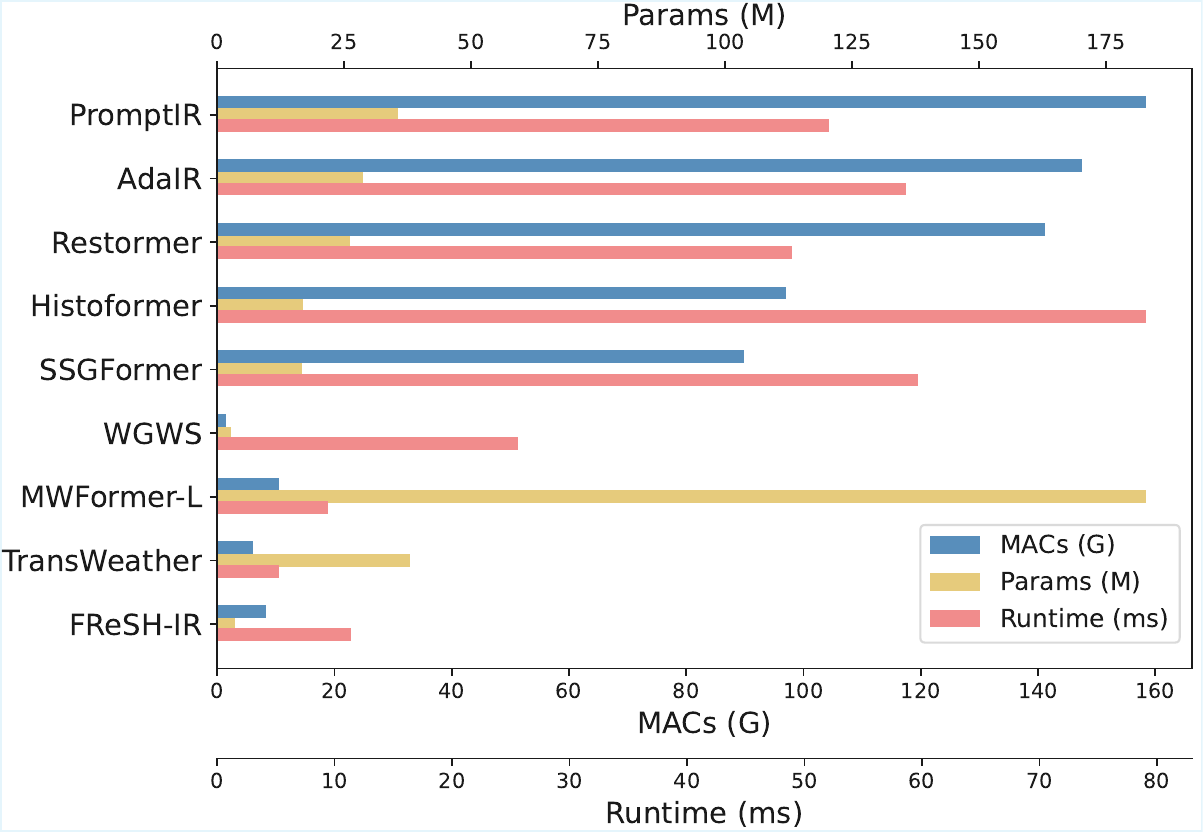}
  \caption{\textbf{Efficiency analysis} across models. TransWeather is the only comparable method to FReSHIR's efficiency. However, we improve its average performance by $+1.67$dB and achieve better perceptual metrics on real-world data.}
  \label{fig:efficiency}
\end{figure}

FReSH-IR is designed for efficient deployment on edge devices. Although performance is sometimes lower, the efficiency of the proposed network is critical for on-device applications. As shown in \Cref{tab:allweather}, only MWFormer-L, WGWS, and TransWeather have comparable efficiency, yet each falls short in some indicator: MWFormer-L~\cite{zhu2024mwformer}'s 182.81M parameters, or WGWS~\cite{zhu2023Weather}'s doubling our runtime. \Cref{fig:efficiency} illustrates this comparison. Our design proves efficient in all indicators, making it suitable for edge device deployment.  A study on inference times at different resolutions is included in the Supplementary Material.

\subsection{Ablation Studies}
To demonstrate that the FReSH-IR architecture is constructed using components that exhibit optimal performance, we conducted a series of ablation studies on various components, as presented in \Cref{tab:ablations_structure}. These ablations provide empirical evidence supporting the appropriateness of the preliminaries outlined in \Cref{subsec:preliminaries}, which informed the network's design. First, by reversing the cutoff of the Fourier masks to process higher frequency information at deeper levels, we observed a decline in network performance. This finding indicates that processing HF with downsampled features is not an effective method for image restoration. Instead, our network design prioritizes the restoration of HF at the upper levels of the decoder, as illustrated in the architecture in \Cref{fig:arch}. 

We also investigated the effects of the encoder restoring both LF and HF. Our findings indicate that this approach does not yield any improvement, thereby reinforcing our strategy of initially restoring LF in the encoder and HF in the decoder. To demonstrate the efficacy of the dual attention module, we replaced this component with a standard NAFBlock~\cite{chen2022simple}, which resulted in decreased performance. This outcome further underscores the significance of the GAP operation as discussed in \Cref{subsec:preliminaries}. 

In addition to our investigation of the components, we conducted a study on possible loss functions---see \cref{tab:ablations_losses}.
We provide additional studies in the Supp. Material.

\begin{table}[t]
    \centering
    \begin{adjustbox}{width=0.48\textwidth}
    \begin{tabular}{p{5cm}ccc}
    \toprule
    Ablation & PSNR$\uparrow$ & SSIM$\uparrow$ & LPIPS$\downarrow$ \\
    \midrule
    1) ResBlocks process high and low & 24.62 & 0.790 & 0.234 \\
    2) NAFBlock instead of DAM & 27.53 & 0.857 & 0.157 \\
    3) Learnable cutoffs & 27.43 & 0.867 & 0.144 \\
    4) Inverted cutoff values & 27.61 & 0.865 & 0.148 \\        
    \midrule
    FReSH-IR & 28.10 & 0.871 & 0.140 \\
    \bottomrule
    \end{tabular}
    \end{adjustbox}
    \caption{\textbf{Ablation study on the components} of FReSH-IR.}
    \label{tab:ablations_structure}
\end{table}

\begin{table}[t]
    \centering
    \begin{adjustbox}{max width=\linewidth}
    \begin{tabular}{cccc}
    \toprule
    Loss & PSNR$\uparrow$ & SSIM$\uparrow$ & LPIPS$\downarrow$ \\
    \midrule
    $\mathcal{L}_{px}$ & 27.48 & 0.862 & 0.155 \\
    $\mathcal{L}_{px} + \mathcal{L}_{freq}$ & 28.07 & 0.871 & 0.142 \\
    \midrule
    $\mathcal{L}_{px} + \mathcal{L}_{enh} + \mathcal{L}_{freq}$ & 28.10 & 0.871 & 0.140 \\
    \bottomrule
    \end{tabular}
    \end{adjustbox}
    \caption{\textbf{Ablation study on training losses}.}
    \label{tab:ablations_losses}
\end{table}

\subsection{Limitations}
The performance of the method on real-world scenes is still limited for demanding downstream tasks, such as autonomous driving. This is common among all the methods and can be attributed to the limited availability of realistic training datasets. Compared to other tasks, such as low-light image enhancement~\cite{yang2021sparse,hai2023r2rnet}, collecting real samples for weather restoration is quite difficult, which remains one of the main challenges in this field. Future work should focus on the creation of these datasets rather than on architectural development. 

\section{Conclusion}

We presented FReSH-IR, an efficient all-in-one weather restoration method designed for edge applications. Through the Spectral Harmonization Block and specialized frequency branches, FReSH-IR decomposes and fuses spectral information at each scale, achieving restoration quality competitive with far heavier models at up to 90\% lower computational cost in terms of memory, operations, and runtime. This trade-off is not a limitation but a deliberate design decision: frequency-aware architecture can recover most of the quality of state-of-the-art methods at a fraction of the resources. We believe this work provides a practical pathway towards real-world, on-device image restoration, and encourages further research into efficiency-first frameworks that treat deployment constraints as a core design objective.

%% file: sec/supplementary.tex
\section{Dataset overview}

\noindent{\textbf{Training dataset.}} Following previous studies \cite{valanarasu2022transweather, li2020all}, we use the \textit{AllWeather} set. It contains a subset of $9,000$ images from Outdoor-Rain \cite{li2019heavy}, which encompasses synthetic images with haze and rain-streaks altogether; $1,069$ images from RaindropData \cite{qian2018attentive}, containing real raindrop images; and $9,000$ images from the Snow100K \cite{liu2018desnownet} training dataset, featuring synthetic images affected by snow.

\vspace{1mm}
\noindent{\textbf{Rain + Fog test set.}} For both deraining and dehazing, we use the Test1 subset from Outdoor-Rain \cite{li2019heavy}. It contains $750$ pairs of synthetic images degraded with rain streaks and haze with different intensities and directions.

\vspace{1mm}
\noindent{\textbf{Raindrop test set.}} The dataset used for testing de-raindrop is the Raindrop A subset from RaindropData \cite{qian2018attentive}. It contains $58$ pairs of real raindrop images with different shapes and intensities.

\vspace{1mm}
\noindent{\textbf{Snow test set.}} For desnowing, we use the Snow100K-L test set from Snow100K \cite{liu2018desnownet}. It contains $16,801$ pairs of synthetic snow images. We also used the Snow100K realistic subset for the visual results. This test set contains $1,329$ images of realistic snow, without their ground truth.

\vspace{1mm}
\noindent{\textbf{Downstream task evaluation.}} To assess the impact of FReSH-IR on downstream perception tasks, we additionally evaluate on two datasets that are not seen during training. Rainy Cityscapes \cite{Hu_2019_CVPR} augments the Cityscapes urban driving scenes with synthetic rain, providing $9,4232$ training and $1,188$ test images with semantic segmentation and object detection ground truth. CSD \cite{chen2021all} is a synthetic snow benchmark containing $8,000$ training and $2,000$ test images.

\section{Frequency analysis.}
\begin{figure*}[t!]
    \centering
    \tabcolsep 8pt
    \begin{tabular}{cc}
\includegraphics[width=0.5\linewidth]{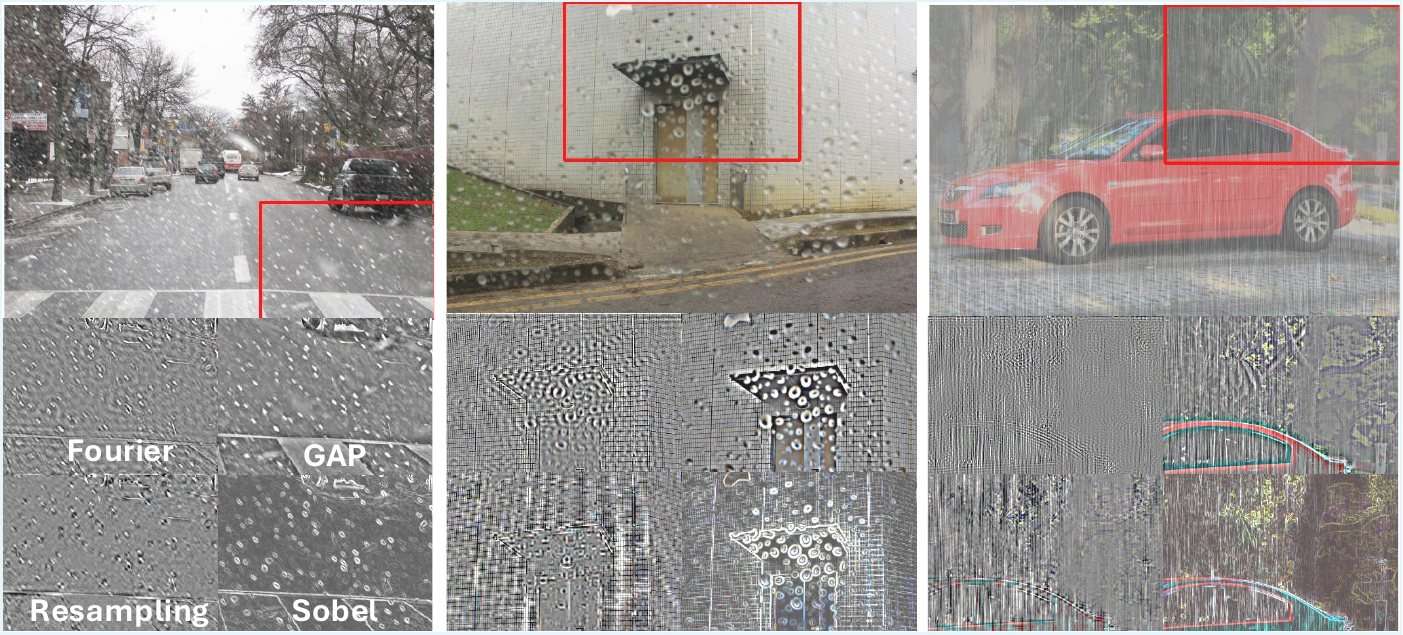} & \includegraphics[width=0.37\linewidth]{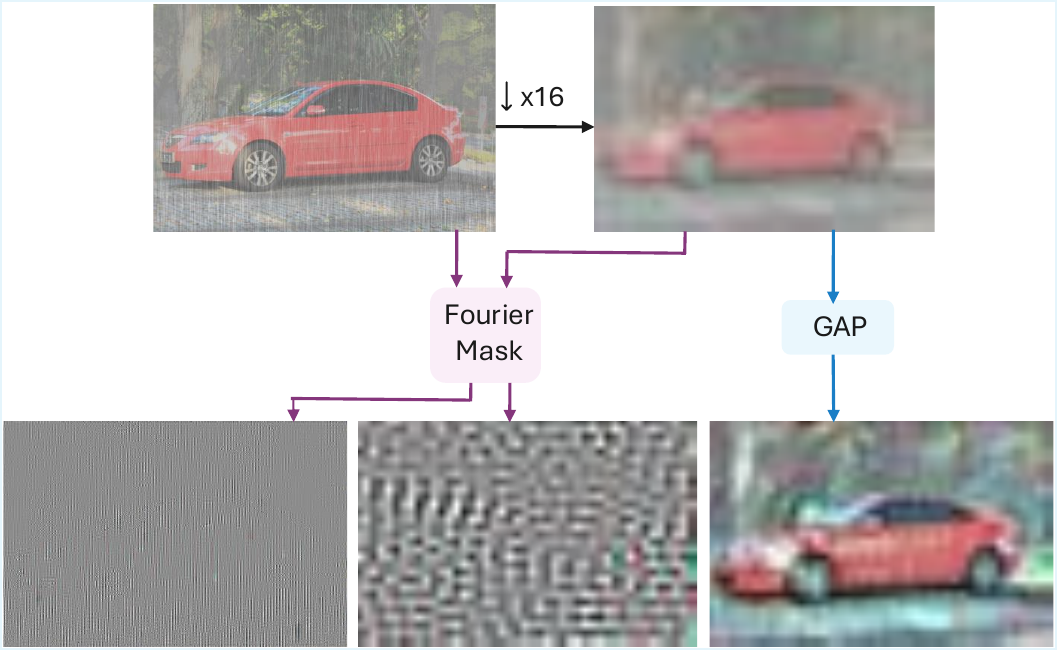}\\
(a) & (b)\\
    \end{tabular}
\caption{\textbf{(a)} Different high-frequency extraction techniques used across three degradations. Top row shows the degraded image with a highlighted region in red. Bottom row shows the extracted high-frequencies (HF) of the indicated region, using four methods. \textbf{(b)} Fourier mask result on the input degraded image and its downsampled version. It also shows the result of applying GAP to the downsampled version of the input. Zoom in for a better view.}
     \label{fig:main_figure}
\end{figure*}

This analysis is motivated by the observation that predominant weather artifacts (\textit{e.g.}, snowflakes, raindrops, etc) mainly appear in the high-frequency domain of degraded images. As illustrated in \Cref{fig:main_figure}~(a), we evaluate each extraction method on the three considered weather degradation types. Our analysis reveals that FFT decomposition yields the most precise isolation of weather artifacts, as it captures degradation-related information while suppressing structural edge content from the underlying clean image. While 2D Global Average Pooling (GAP)--simple mean kernel-- also demonstrates strong degradation extraction capability, it preserves some clean image edges, which helps maintain fine structural details during the restoration process. Although Sobel filtering captures both degradation patterns and structural details similarly to GAP, the weather artifacts are less visible while producing more pronounced edge responses, resulting in a less favorable trade-off for degradation isolation. Finally, the resampling-based approach fails to provide meaningful information, making it unsuitable for this purpose. 

However, beyond a certain depth, Fourier decomposition loses the ability to capture meaningful information altogether, as shown in \Cref{fig:main_figure}~(b). This is the moment when we transition to GAP-based extraction, which, despite operating on deep feature representations, retains the ability to capture structural image edges, as illustrated in \Cref{fig:main_figure}~(b).

\section{Additional efficiency results.}
To remark the efficiency potential of our proposal, we include the image runtime study conducted on a 24GB RTX 4090 GPU. We conducted the same study on a 16GB Jetson Orin NX. Both experiments are illustrated on \Cref{tab:runtime_sizes}.

\begin{table*}[!ht]
  \centering
  \footnotesize
  \setlength{\tabcolsep}{5pt}
  \begin{tabular}{lccccc|cccc}
    \toprule
    & \multicolumn{5}{c|}{RTX 4090} & \multicolumn{4}{c}{Jetson Orin NX} \\
    \cmidrule(lr){2-6} \cmidrule(lr){7-10}
    Method & 360p & 720p & 1080p & 1440p & 2160p & 360p & 720p & 1080p & 1440p \\
    \midrule
    SSGFormer   & 0.400 & OOM & OOM & OOM & OOM & 4.413 & OOM & OOM & OOM \\
    AdaIR       & 0.324 & 1.294 & 3.154 & OOM & OOM & 2.632 & 11.148 & OOM & OOM \\
    HistoFormer & 0.453 & 1.793 & 4.385 & OOM & OOM & 4.429 & 19.371 & OOM & OOM \\
    \midrule
    \rowcolor{lyellow} WGWS & 0.094 & 0.363 & 0.787 & \underline{1.369} & \underline{3.051} & 0.863 & 3.371 & 8.057 & \underline{16.600} \\
    \rowcolor{lyellow} MWFormer & \underline{0.029} & \underline{0.155} & 0.486 & OOM & OOM & \underline{0.198} & \underline{1.360} & OOM & OOM \\
    \rowcolor{lyellow} TransWeather & \textbf{0.008} & \textbf{0.053} & \textbf{0.190} & OOM & OOM & \textbf{0.085} & \textbf{0.616} & \textbf{3.570} & OOM \\
    \midrule
    \textbf{FReSH-IR} & 0.045 & 0.178 & \underline{0.394} & \textbf{0.690} & \textbf{1.595} & 0.458 & 1.831 & \underline{4.329} & \textbf{7.913} \\
    \bottomrule
  \end{tabular}
  \caption{Runtime in seconds for different image resolutions, averaged over 100 frames, on different devices. \textbf{Bold} and \underline{underlined} represent the best and second-best results, respectively.}
  \label{tab:runtime_sizes}
\end{table*}

\begin{table*}[!ht]
  \centering
  \begin{minipage}[t]{0.56\textwidth}
    \centering
    \begin{adjustbox}{max width=\textwidth}
    \begin{tabular}{lcccc}
      \toprule
      Ablation & Cost & PSNR$\uparrow$ & SSIM$\uparrow$ & LPIPS$\downarrow$ \\
      \midrule
      $\cdot$Encoder processes only highs & 3.46/8.32 & 27.27 & 0.859 & 0.153 \\
      $\cdot$NAFBlock in the Decoder & 4.02/9.05 & 28.20 & 0.872 & 0.139 \\
      $\cdot$Restormer block on decoder highs & 3.44/8.21 & 25.13 & 0.794 & 0.233 \\
      $\cdot$Decoder does not process lows & 3.42/8.06 & 25.18 & 0.814 & 0.193 \\
      $\cdot$SHB outputs concatenated freqs & 3.65/8.93 & 23.39 & 0.705 & 0.295 \\
      $\cdot$Same cutoff value & 3.46/8.32 & 27.86 & 0.868 & 0.144 \\
      $\cdot$Traditional skip-connections & 3.46/8.32 & 28.06 & 0.871 & 0.141 \\
      \midrule
      \textbf{FReSH-IR} & 3.46/8.32 & 28.10 & 0.871 & 0.140 \\
      \bottomrule
    \end{tabular}
    \end{adjustbox}
  \end{minipage}
  \hfill
  \begin{minipage}[t]{0.40\textwidth}
    \centering
    \begin{adjustbox}{max width=\textwidth}
    \begin{tabular}{ccccc}
      \toprule
      Channels & Cost & PSNR$\uparrow$ & SSIM$\uparrow$ & LPIPS$\downarrow$ \\
      \midrule
      16 & 1.57/3.90/11.13 & 25.20 & 0.814 & 0.217 \\
      48 & 13.47/31.44/13.40 & 26.75 & 0.848 & 0.168 \\
      \midrule
      24 & 3.46/8.32/11.42 & 28.10 & 0.871 & 0.140 \\
      \bottomrule
    \end{tabular}
    \end{adjustbox}
  \end{minipage}
  \caption{\textbf{(a)} Ablation study of the components of FReSH-IR. Cost: Params (M) / MACs (G).
  \textbf{(b)} Ablation study on embedding channels. Cost: Params (M) / MACs (G) / Runtime (ms) on RTX 4090.}
  \label{tab:supp_ablations}
\end{table*}

We conclude that (i) only FReSH-IR and WGWS~\cite{zhu2023Weather} handle resolutions of at least 1440p, and (ii) FReSH-IR demonstrates superior scalability in inference times relative to image resolution compared to other methods. The alternative methods encounter Out-of-Memory (OOM) errors.
  
\section{Additional ablation studies}
To further justify the selection of our FReSH-IR architecture, we conducted additional experimental studies. The results of these studies are presented in \Cref{tab:supp_ablations} (a). We attempted to process only the high frequencies on the encoder, but the results showed a significant decrease in performance, demonstrating that restoring low frequencies in the encoder is more effective.  

We also performed experiments on the decoder block. First, we replaced our proposed FTB with separate NAFBlocks for each frequency component. This study showed only a negligible improvement in performance while increasing the computational cost, which reinforces our idea of using specialized branches for the high- and low-frequency components. When processing the high-frequency components with the Restormer Transformer, cost was slightly lower while the restoration performance significantly dropped, demonstrating the potential of our approach. We also tried processing only the high frequencies on the decoder, which resulted in a significant decrease in performance, further supporting our strategy of restoring both components in the decoder. 

Additionally, we conducted experiments on the SHB. We changed the skip fusion method from addition to concatenation, slightly increasing the cost but severely degrading performance. 

To further support our decision on the cutoff values, we performed an experiment setting the same value for every depth. This experiment showcases the use of different masks in each level, improving the restoration performance.

Moreover, to demonstrate the capabilities of our novel skip-connection mechanism, we substituted our approach with traditional skip-connections.

Finally, we performed experiments regarding the number of channels. The results of this study, displayed in \Cref{tab:supp_ablations} (b), demonstrate that our configuration of 24 channels achieves an optimal balance between performance and computational efficiency. 

\section{Qualitative results}
\begin{figure*}[p]
    \centering
    \begin{tabular}{c}
         \includegraphics[width=\textwidth]{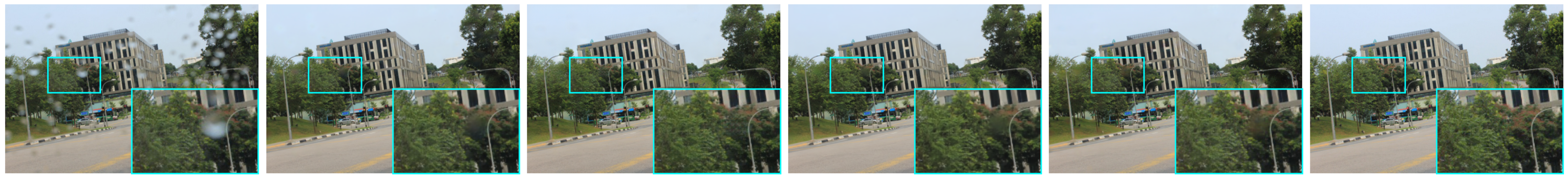} \\
         \includegraphics[width=\textwidth]{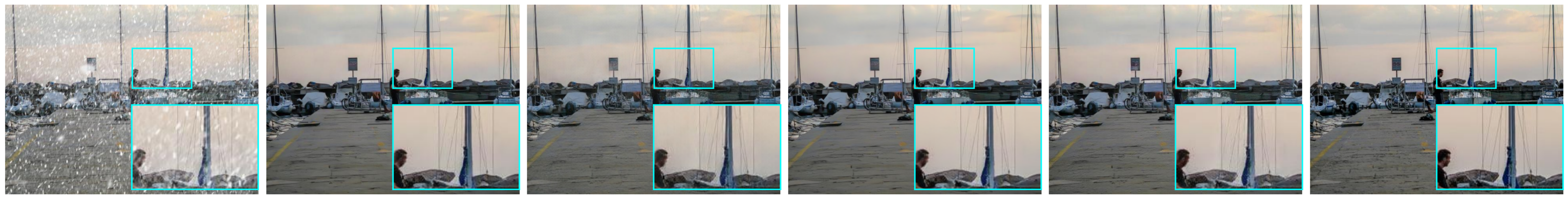} \\
         \includegraphics[width=\textwidth]{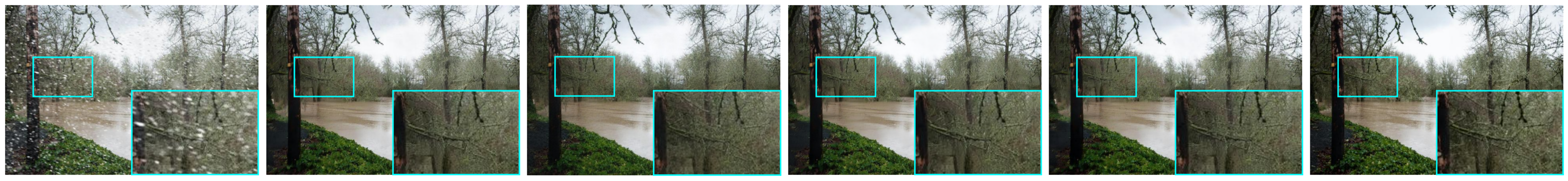}\\
         \includegraphics[width=\textwidth]{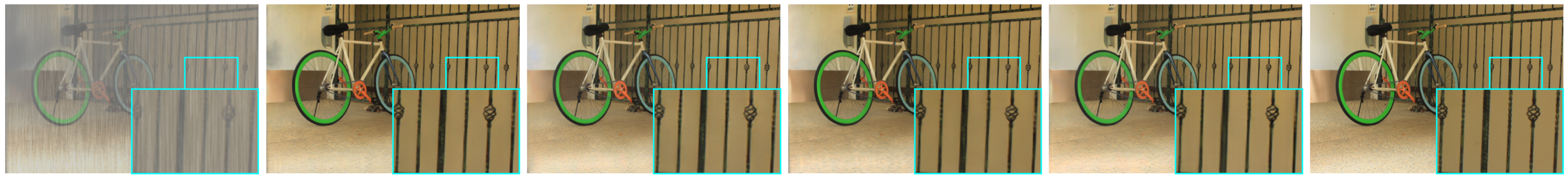}\\
         \includegraphics[width=\textwidth]{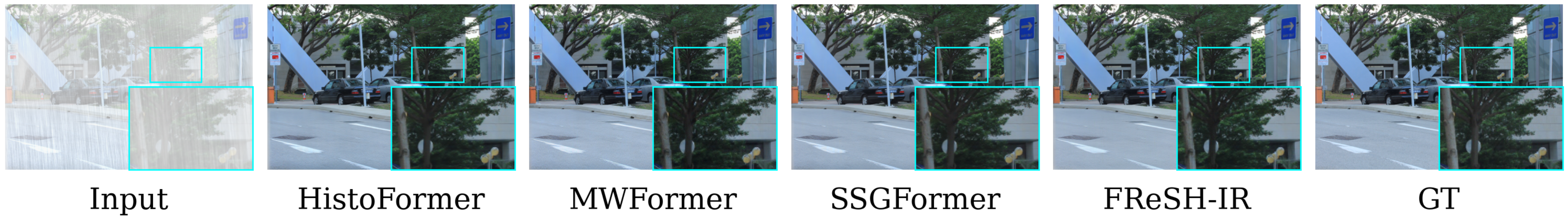} \\
    \end{tabular}
    \caption{Additional qualitative results obtained using our FReSH-IR. From top to bottom: RainDrop, Snow100K, OutdoorRain datasets. Our method achieves similar results to state-of-the-art methods at a fraction of their computational cost.}
    \label{fig:supp_qualis}
\end{figure*}

\begin{figure*}[p]
    \centering
    \begin{tabular}{c}
         \includegraphics[width=\textwidth]{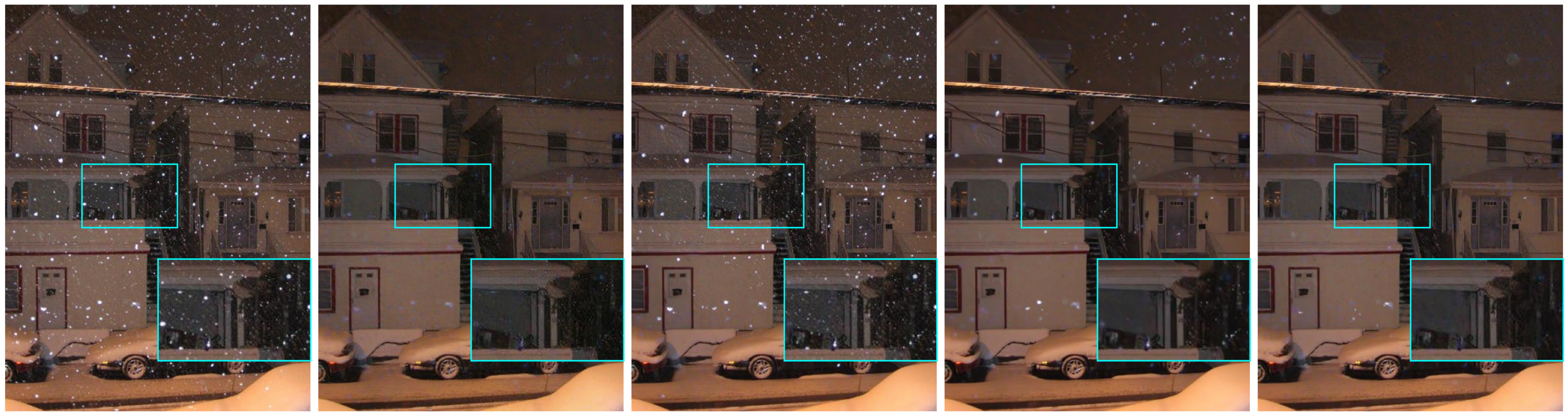} \\
         \includegraphics[width=\textwidth]{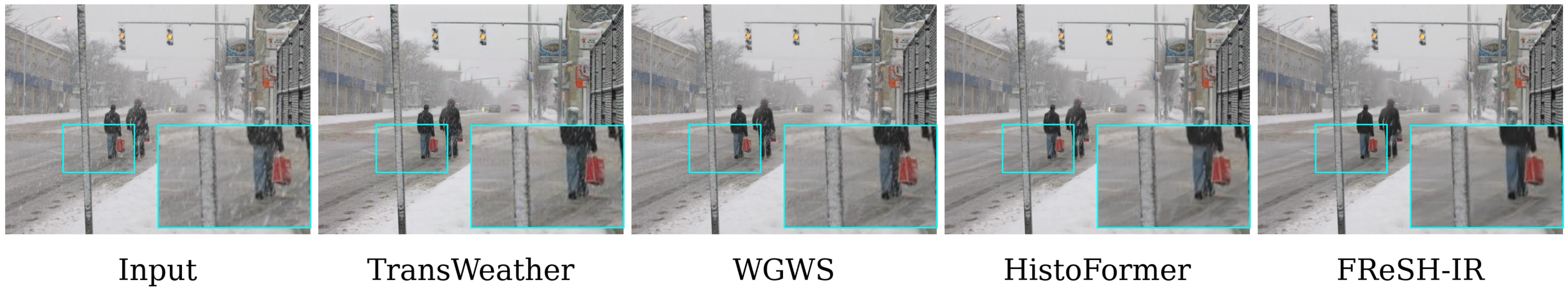} \\
    \end{tabular}
    \caption{Additional qualitative results from realistic Snow100k.}
    \label{fig:supp_real}
\end{figure*}

In \Cref{fig:supp_qualis} we provide a comparison of visual results in synthetic and real benchmarks.
\Cref{fig:supp_real} shows additional qualitative results on the realistic subset of real Snow100K against several methods. 

